\documentclass[lettersize,journal]{IEEEtran}
\usepackage{amsmath,amsfonts}
\usepackage{float}
\usepackage{algorithm}
\usepackage{algpseudocode}
\usepackage{array}
\usepackage[caption=false,font=normalsize,labelfont=sf,textfont=sf]{subfig}
\usepackage{textcomp}
\usepackage{stfloats}
\usepackage{url}
\usepackage{verbatim}
\usepackage{graphicx}
\def\BibTeX{{\rm B\kern-.05em{\sc i\kern-.025em b}\kern-.08em
    T\kern-.1667em\lower.7ex\hbox{E}\kern-.125emX}}
\usepackage{balance}

\usepackage{hyperref}
\usepackage{booktabs}
\usepackage{multirow}
\usepackage{makecell}
\begin{document}

\title{A Parallel Hybrid Action Space Reinforcement Learning Model for Real-world Adaptive Traffic Signal Control}

\author{
\IEEEauthorblockN{
Yuxuan Wang\IEEEauthorrefmark{1}, 
Meng Long\IEEEauthorrefmark{1}, 
Qiang Wu\IEEEauthorrefmark{2}, 
Wei Liu\IEEEauthorrefmark{3},
Jiatian Pi\IEEEauthorrefmark{1}\textsuperscript{\S}, 
and Xinmin Yang\IEEEauthorrefmark{1}
} 

\IEEEauthorblockA{\IEEEauthorrefmark{1}National Center For Applied Mathematics In Chongqing \\ Chongqing Normal University, Chongqing 401331, China\\ 
Email: wangyuxuan@stu.cqnu.edu.cn, longmeng@cqnu.edu.cn, pijiatian@cqnu.edu.cn, xmyang@cqnu.edu.cn} 
\IEEEauthorblockA{\IEEEauthorrefmark{2}Institute of Fundamental and Frontier Sciences\\ University of Electronic Science and Technology of China, Chengdu 611731, China\\ Email: qiang.wu@uestc.edu.cn}
\IEEEauthorblockA{\IEEEauthorrefmark{3}Chongqing Jiaotong University, Chongqing 400074, China\\ Email: neway@cqjtu.edu.cn}

% # -----------------未修改-----------------------
\thanks{\textsuperscript{\S}Corresponding author: Jiatian Pi. Email: pijiatian@cqnu.edu.cn} % 标记通讯作者
\thanks{Sponsored by Natural Science Foundation of Chongqing (Grant No. CSTB2022NSCQ-LZX0040, CSTB2023NSCQ-LZX0012, CSTB2023NSCQ-LZX0160).} % 添加资助项目
% \thanks{Manuscript received January 20, 2002; revised August 26, 2015. This work was supported by the IEEE.}% 
% \thanks{M. Shell was with the Georgia Institute of Technology.}
% \thanks{Manuscript created October, 2020; This work was developed by the IEEE Publication Technology Department. This work is distributed under the \LaTeX \ Project Public License (LPPL) ( http://www.latex-project.org/ ) version 1.3. A copy of the LPPL, version 1.3, is included in the base \LaTeX \ documentation of all distributions of \LaTeX \ released 2003/12/01 or later. The opinions expressed here are entirely that of the author. No warranty is expressed or implied. User assumes all risk.}
% # -----------------未修改-----------------------
}

\markboth{Journal of \LaTeX\ Class Files,~Vol.~18, No.~9, September~2020}%
{How to Use the IEEEtran \LaTeX \ Templates}

\maketitle

\begin{abstract}
Adaptive traffic signal control (ATSC) can effectively reduce vehicle travel times by dynamically adjusting signal timings but poses a critical challenge in real-world scenarios due to the complexity of real-time decision-making in dynamic and uncertain traffic conditions. The burgeoning field of intelligent transportation systems, bolstered by artificial intelligence techniques and extensive data availability, offers new prospects for the implementation of ATSC. In this study, we introduce a parallel hybrid action space reinforcement learning model (PH-DDPG) that optimizes traffic signal phase and duration of traffic signals simultaneously, eliminating the need for sequential decision-making seen in traditional two-stage models. Our model features a task-specific parallel hybrid action space tailored for adaptive traffic control, which directly outputs discrete phase selections and their associated continuous duration parameters concurrently, thereby inherently addressing dynamic traffic adaptation through unified parametric optimization. %Our model features a unique parallel hybrid action space that allows for the simultaneous output of each action and its optimal parameters, streamlining the decision-making process. 
Furthermore, to ascertain the robustness and effectiveness of this approach, we executed ablation studies focusing on the utilization of a random action parameter mask within the critic network, which decouples the parameter space for individual actions, facilitating the use of preferable parameters for each action. The results from these studies confirm the efficacy of this method, distinctly enhancing real-world applicability. We implemented PH-DDPG with both offline and online frameworks to provide deployment flexibility. Evaluations carried out across multiple real-world datasets, and the introduction of two innovative metrics validate the superior performance of PH-DDPG over existing state-of-the-art methods. Our code is released on \href{https://github.com/cjwyx/PH-DDPG-model}{Github}.
\end{abstract}


\section{Introduction}

\begin{figure}[htbp]
	\centering
	\includegraphics[width=0.46 \textwidth]{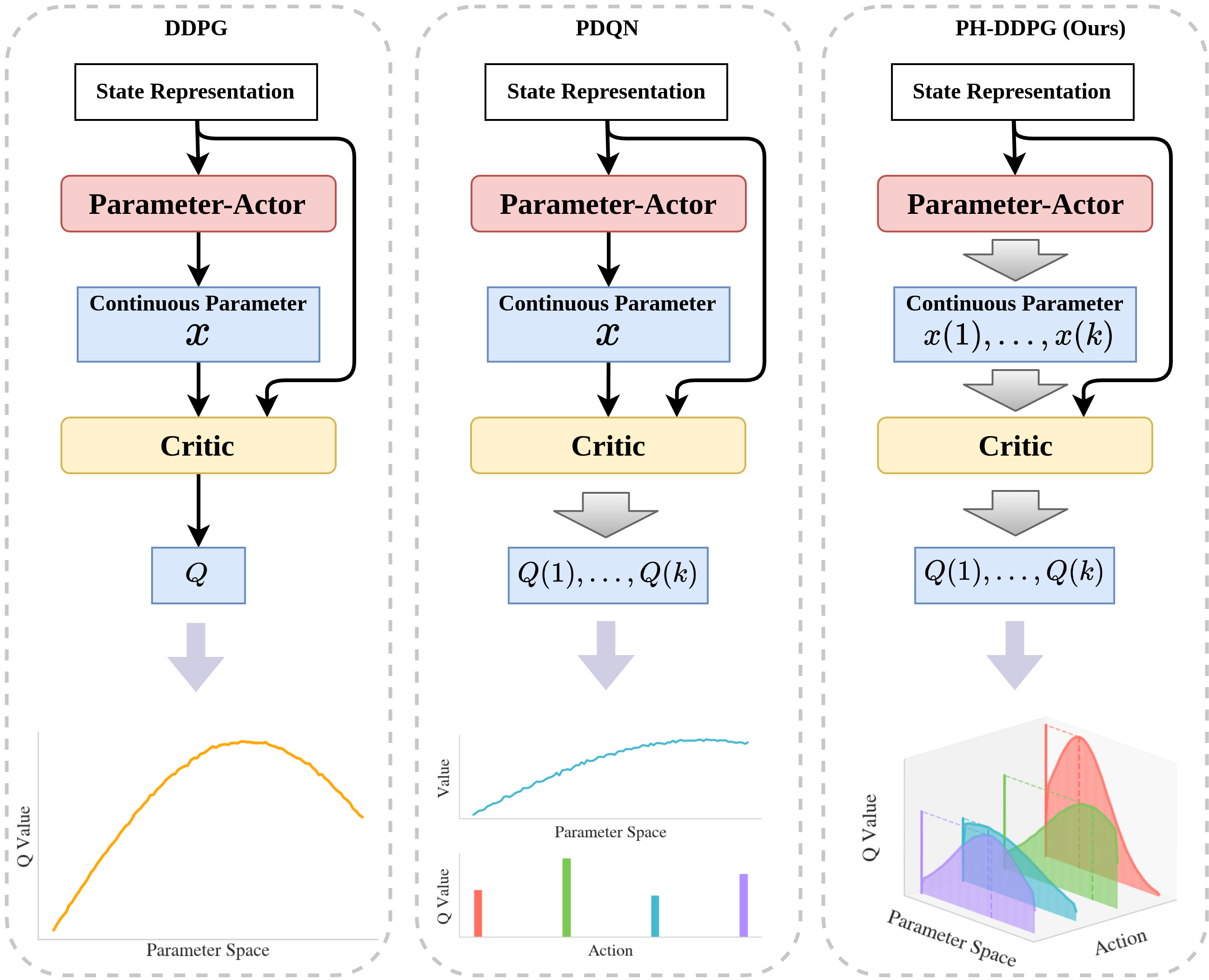}
	\caption{This figure illustrates the structural differences in reinforcement learning among DDPG, P-DQN, and PH-DDPG. 
	DDPG outputs a single continuous action by maximizing the Q-value for the optimal action parameter. 
	P-DQN focuses on selecting the optimal discrete action under a fixed optimal action parameter. 
	PH-DDPG extends these approaches by computing the Q-value for each discrete action under its respective optimal action parameter. As a result, PH-DDPG generates multiple continuous action parameters in the Actor part, distinguishing it from the other two methods and enabling more flexible and adaptive decision-making in hybrid action spaces.}
	\label{introduction1}
\end{figure}

\IEEEPARstart{T}{he} optimization of traffic signal control (TSC) is critical for alleviating urban congestion and enhancing road safety, which is a challenging real-world problem and essential for an efficient transportation system. Traditionally, TSC heavily rely on predefined rules by transportation engineers and field calibration, which are not flexible enough to handle highly dynamic traffic scenarios \cite{hunt1982scoot}. Until now, predetermined traffic signal plans are still widely used in a substantial portion of practical intersections \cite{tang2019global}. However, as well as the increasing number of vehicles, the diversified driving behaviors such as navigating on autopilot \cite{guo2019urban}, and the urgent demand for reducing energy consumption. It is becoming much more significant to develop highly adaptive traffic signal control (ATSC), which aims to dynamically adjust signal timing according to real-time traffic conditions. 

ATSC approaches have not only attracted considerable attention from research communities, but have also gained wide recognition for their effectiveness and efficiency in practical implementation. Since the 1990s, a large amount of work has been proposed to optimize the ATSC system, such as model-based methods \cite{daganzo1995cell,ye2019survey}, max pressure-based methods \cite{varaiya2013max}, and data-driven approaches \cite{wei2021recent}. In recent years, reinforcement learning (RL) techniques have emerged as a popular approach for ATSC and have achieved impressive performance in traffic simulation environments. The RL-based methods can learn of an optimised policy from the traffic environment via trial and error without relying on predefined rules. The appealing nature of this property has long attracted researchers to utilize RL in addressing traffic signal control issues. Initially, SARSA \cite{thorpe1996tra} and tabular Q-learning \cite{abdulhai2003reinforcement} are used to solve highly simplified ATSC problems, where the states are required to be discrete and low-dimensional. With the development of deep reinforcement learning \cite{mnih2015human}, many deep RL-based ATSC approaches such as IntelliLight \cite{wei2018intellilight}, FRAP \cite{zheng2019learning}, PressLight \cite{wei2019presslight}, AttendLight \cite{oroojlooy2020attendlight}, Efficient-XLight \cite{wu2021efficient}, DynamicLight \cite{zhang2022dynamiclight}, DuaLight \cite{lu2023dualight}, leverage highly expressive deep neural networks to effectively process more complex and high-dimensional states. The impressive results achieved by them demonstrate the superiority of deep RL data-driven approaches over conventional methods. Othter advanced modeling techniques have also been adopted to deep RL-based ATSC methods, which improve the signal control scalability from a intersection to thousands of lights in a large-scale traffic network \cite{wei2019colight,chen2020toward,lin2023denselight}. 

Despite the impressive performance of these methods in simulation environments, their real-world deployment remains limited due to the complexities of practical implementation and the challenges of sim-to-real transfer. Most RL-based ATSC methods use fixed action durations, which fail to achieve dynamic phase durations based on real-time traffic conditions. Authors typically add or subtract a fixed five seconds to adjust the current duration in the next cycle \cite{liang2019deep,wei2019colight,oroojlooy2020attendlight,zhang2022expression}. Unfortunately, such fixed action durations can significantly influence model performance \cite{zhang2022expression} and limit the capability of RL-based methods. CycLight \cite{han2024cyclight} adopts a novel cycle-level strategy to simultaneously optimize cycle length and splits using Parameterized Deep Q-Networks (P-DQN). By jointly evolving both discrete actions and continuous parameters, CycLight enhances the flexibility of phase durations. However, it remains heavily dependent on a cyclical phase structure, which does not fully support real-time adaptive signal control.

Establishing the relationship between phase selection and duration computation in a non-cyclical structure is crucial for RL-based model training. Specifically, phase selection is determined by discrete actions, whereas duration can be treated as a continuous action. Both DynamicLight\cite{zhang2022dynamiclight} and H-PPO\cite{HPPO2024} adhere to the Centralized Training Decentralized Execution (CTDE) framework, facilitating asynchronous decision-making for traffic signal control. Researchers often divide the ATSC process into two stages, as seen in DynamicLight, which separates the optimization of phase control and duration control. DynamicLight, based on Dueling DQN, uses a single agent to predict runtime, with two models providing discrete actions and time parameters. In contrast, H-PPO leverages multiple agents and extends the PPO algorithm by incorporating multiple policy heads to approximate parameterized actions. This approach allows one model to provide discrete actions, while another model optimizes the parameters based on the selected discrete actions, enabling direct optimization in the hybrid action space. Despite their advancements, both models face challenges in hybrid action space decision-making, particularly due to the two-step decision process where the evaluation of discrete actions does not consider action parameters.

% Despite their advancements, both models face challenges in hybrid action space decision-making. Specifically, the two-step decision process—where discrete action selection does not account for action parameters—often leads to convergence to local optima. Therefore, it is essential to develop methods that integrate action evaluation with parameter conditions to achieve more optimal results.

To address these challenges, we develop a \textbf{Parallel Hybrid Action Space method based on Deep Deterministic Policy Gradient for real-world adaptive traffic signal control (PH-DDPG)}. As shown in Figure \ref{introduction1}, PH-DDPG assesses each reward based on its corresponding action parameter, yet completes the overall optimization collaboratively. Our approach leverages the concept of \textbf{disaggregated rewards}, first proposed in \cite{Casas2017DeepDP}, which was a pioneering work in applying DDPG to Traffic Signal Control (TSC). The disaggregated rewards concept allows for more granular evaluation of actions, leading to better optimization. By further advancing this concept, PH-DDPG innovatively uses a single critic to evaluate each reward independently for its respective action parameters while achieving joint optimization.

The main contributions of this paper are summarized as follows:

\begin{enumerate}
\item{We design a method that can optimize multiple hybrid action spaces in parallel, enabling the selection of actions with better action parameters. This ensures the use of optimal phases with the best parameters in signal control processes.}

\item{We evaluate the effectiveness of our proposed method in seven scenarios. The experimental results demonstrate that our model achieves state-of-the-art performance in single-agent environments. Additionally, we create several variants of our model to adapt to different application needs and conduct comparative studies.}

\item{We provide an in-depth study of our proposed method for decoupling multiple hybrid action spaces, including mathematical proofs of parallel optimization and tests of the decoupling effectiveness under various configurations.}
\end{enumerate}  

The remainder of the paper is structured as follows. Section 2 (Related Work) reviews existing literature on Deep Reinforcement Learning (DRL) for Traffic Signal Control (TSC) and Hybrid Action Space Reinforcement Learning, identifying gaps and challenges. Section 3 (Problem Definition) formulates the reinforcement learning problem and defines the traffic environment configuration. Section 4 (Methodology) presents the PH-DDPG approach, detailing action feature representations, neighboring nodes feature embedding, and the decoupled parameterized action reinforcement learning framework, along with implementation specifics. Section 5 (Performance Evaluation) describes the real-world datasets, experimental settings, comparative methods, and introduces new evaluation metrics. Section 6 (Performance Comparison) evaluates the overall performance of PH-DDPG, its variants in offline mode, conducts an ablation study, and examines hyperparameter sensitivity

\section{Related Work}
\subsection{Deep Reinforcement Learning for TSC}

Deep Reinforcement Learning (DRL) has significantly advanced Traffic Signal Control (TSC) by leveraging highly expressive deep neural networks to process complex and high-dimensional states. Early approaches like SARSA and tabular Q-learning, introduced by \cite{thorpe1996tra} and \cite{abdulhai2003reinforcement} respectively, laid the groundwork by addressing simplified TSC problems with discrete state representations. However, the development of DRL has enabled more sophisticated methods such as IntelliLight \cite{wei2018intellilight}, FRAP \cite{zheng2019learning}, PressLight \cite{wei2019presslight}, AttendLight \cite{oroojlooy2020attendlight}, Efficient-XLight \cite{wu2021efficient}, DynamicLight \cite{zhang2022dynamiclight}, and DuaLight \cite{lu2023dualight}, which utilize deep learning to optimize traffic signal timings dynamically. These methods have demonstrated superior performance in traffic simulations, showcasing the potential of DRL in TSC. 
Additionally, hybrid and advanced modeling techniques have been explored to address the limitations of fixed action durations. For instance, CycLight \cite{han2024cyclight} adopts a cycle-level strategy using Parameterized Deep Q-Networks (P-DQN) to optimize both cycle length and splits, enhancing phase duration flexibility. Other approaches like CoLight \cite{wei2019colight} and DenseLight \cite{lin2023denselight} improve scalability from single intersections to large-scale traffic networks. Despite these advancements, real-world deployment of DRL-based TSC methods remains limited due to practical implementation complexities and sim-to-real transfer challenges. Most RL-based TSC methods still relies on fixed action durations, which can significantly influence model performance and limit adaptability to real-time traffic conditions. Future research should focus on developing more dynamic and adaptive DRL models that can handle real-time traffic variations effectively, integrating fairness criteria as seen in the FSPPO algorithm to ensure balanced waiting times for all drivers. In summary, while DRL has shown remarkable progress in TSC, overcoming real-world deployment challenges and enhancing model adaptability remain critical areas for future research.

\subsection{Hybrid Action Space Reinforcement Learning}

Hybrid Action Space Reinforcement Learning has emerged as a promising approach to address the limitations of traditional discrete or continuous action spaces by combining both. Q-PAMDP focuses on learning how to choose an action with continuous parameters by alternately training the action selection process and the continuous parameter selection process, using Q-PAMDP(1) and Q-PAMDP($\infty$) to indicate different update frequencies of the policy network \cite{masson2016reinforcement}. PA-DDPG extends the Deep Deterministic Policy Gradient (DDPG) algorithm to handle hybrid action spaces by relaxing the action space into a continuous set and restricting the gradient bounds of the action space through methods like zeroing, squashing, and inverting gradients \cite{Casas2017DeepDP}. MP-DQN addresses the problem of excessive parameterization in P-DQN by using parallel batch processing to assign action parameters to the Q network, thereby reducing the influence of a single discrete action on other continuous action parameters \cite{zhang2021attentionlight}. H-PPO proposes an actor-critic structure in a hybrid action space based on the Proximal Policy Optimization (PPO) architecture, featuring multiple parallel sub-actor networks to process discrete and continuous actions simultaneously, and a global critic network to update policies \cite{HPPO2024}. These methods collectively enhance the flexibility and effectiveness of reinforcement learning in environments requiring both discrete and continuous actions, demonstrating significant improvements in various applications.

\section{Problem Definition}

\subsection{Traffic Environment Configuration}

\begin{figure*}[htbp]
	\centering
	\includegraphics[width=0.98\textwidth]{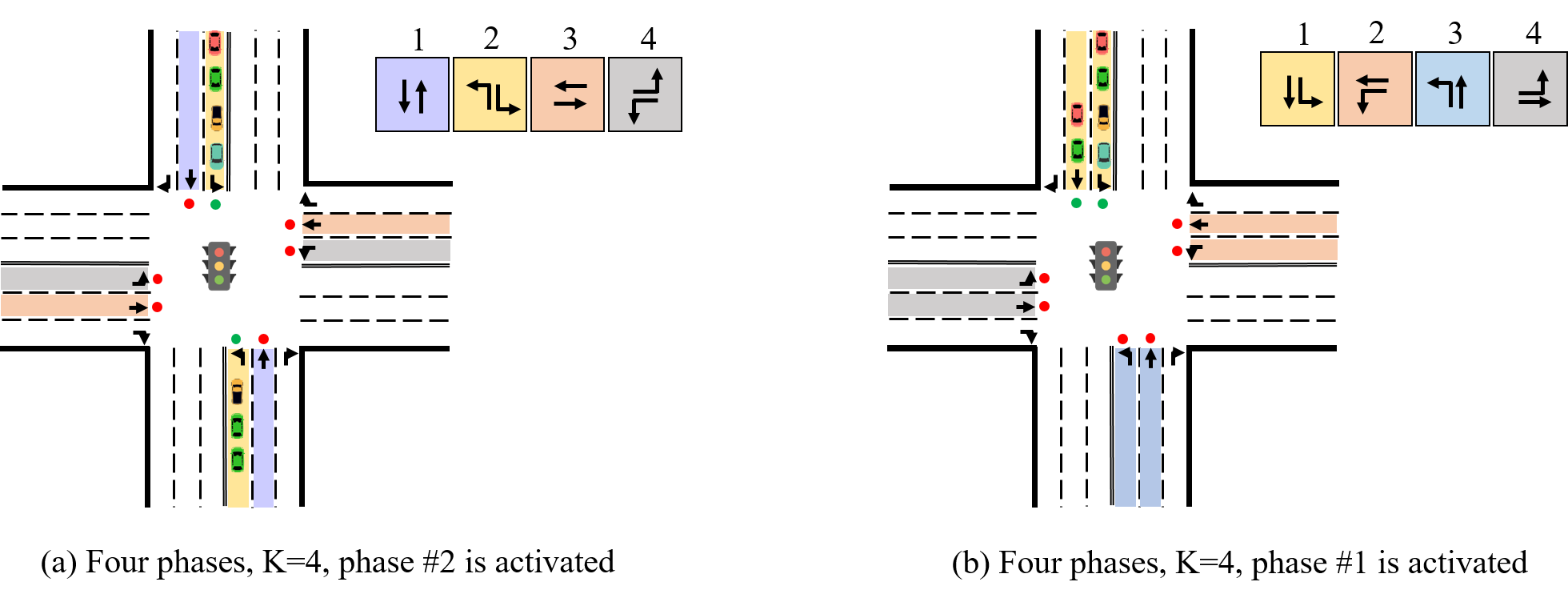}
	\caption{The illustration of an intersection with two different four-phase action set case.}
	\label{figphaseig1}
\end{figure*}

\textbf{Intersection Layout}\quad The traffic environment is structured around intersections and the road network. Each intersection has four entry approaches: North, South, East, and West. Each approach consists of three lanes designated for left-turn, through, and right-turn movements. To manage traffic flow and ensure safety, a three-second yellow signal is implemented at the end of each stage, with a one-second red clearance to prevent collisions during signal transitions. The traffic network is represented as a directed graph where intersections are nodes and roads are edges. Each road comprises several lanes, which are the fundamental units for vehicle movement.

\textbf{Traffic Signal Control}\quad Traffic movements refer to the flow of vehicles in specific directions at intersections, such as left-turn and straight movements. Right-turn movements are generally allowed under any circumstances but must yield to higher-priority movements. Traffic signal control is essential for managing these movements through intersections and road networks. It involves changing signal phases to allow a combination of non-conflicting traffic movements to proceed simultaneously. This control system aims to optimize the flow of traffic based on real-time conditions, ensuring effective and secure vehicle movement.

% \textbf{Traffic network and traffic movement}\quad A traffic network is defined as a directed graph in which intersections and roads are represented by nodes and edges, respectively. Each road consists of several lanes, which are the basic units to support vehicle movement. A traffic movement is defined as traffic moving across the intersection towards a certain direction, such as left-turn and straight. 

% \textbf{Traffic signal control}\quad The movement of vehicles through an intersection or a road network is referred to as traffic flow. Traffic signal control aims to control traffic flow for effective and secure vehicle movement based on the change of signal phase. 

\textbf{Phase action and phase duration}\quad A phase action is a set of various traffic lights that are activated simultaneously in various lanes to indicate traffic movements. This paper focuses on the main signal phases and vehicles that turn right can pass regardless of the signal. The phased duration is the time that each phase spends within the green signal. In our method, the phase duration is not fixed and can dynamically change with different states. 

\subsection{Reinforcement Learning Formulation}

\textbf{Problem formulation based on Dec-POMDPs}\quad The urban traffic signal control task discussed in this paper involves multiple agents making decisions at their respective intersections with input data corresponding to the state information of those intersections. This problem is formulated by decentralized partially observable Markov processes(Dec-POMDPs).

A decentralized partially observable Markov decision process is defined as a tuple $M = \langle D, S, A, T, O, \mathcal{O}, R, h, b_0 \rangle$, where:
\begin{itemize}
\item $D = \{1, \ldots, n\}$ is the set of $n$ agents.
\item $S$ is a (finite) set of states.
\item $A$ is the set of joint actions.
\item $T$ is the transition probability function.
\item $O$ is the set of joint observations.
\item $\mathcal{O}$ is the observation probability function.
\item $R$ is the immediate reward function.
\item $h$ is the horizon of the problem as mentioned above.
\item $b_0 \in \Delta(S)$, is the initial state distribution at time $t = 0$.
\end{itemize}
The Dec-POMDPs model extends single-agent POMDPs models by considering joint actions and observations. In particular, $A = \times_{i \in D} A_i$ is the set of joint actions. Here, $A_i$ is the set of actions available to agent $i$, which can be different for each agent. At every stage $t$, each agent $i$ takes an action $a_{i,t}$, leading to one joint action $a = \langle a_1, \ldots, a_n \rangle$ at every stage. How this joint action influences the environment is described by the transition function $T$, which specifies $\Pr(s'|s,a)$. In a Dec-POMDPs, agents only know their own individual actions; they do not observe each other’s actions. We will assume that $A_i$ does not depend on the stage or state of the environment (but generalisations that do incorporate such constraints are straightforward to specify). Similar to the set of joint actions, $O = \times_{i \in D} O_i$ is the set of joint observations, where $O_i$ is a set of observations available to agent $i$. Every time step the environment emits one joint observation $o = \langle o_1, \ldots, o_n \rangle$ from which each agent $i$ only observes its own component $o_i$. The observation function $\mathcal{O}$ specifies the probabilities $\Pr(o|a,s')$ of joint observations.

 \textbf{Application in Urban Traffic Signal Control} In the context of urban traffic signal control, the Dec-POMDPs framework is utilized to manage multiple intersections where each agent (traffic signal) operates based on local traffic conditions. The state $S$ for each lane $l$ at an intersection is defined as:
\[
S_l = (q_l, m_l, v_{l,1}, v_{l,2}, v_{l,3}, v_{l,4})
\]
where $q_l$ represents the queue length, $m_l$ denotes the number of moving vehicles, $v_{l,i}$ indicates the vehicle count in the $i$-th 100-meter segment approaching the intersection, up to 400 meters. The overall state $S$ is the aggregation of $S_l$ for all lanes at the intersection. The reward function $R$, aimed at minimizing wait times, aggregates the queue lengths across all lanes, thus incentivizing actions that reduce overall congestion and improve traffic flow.

The action space in this application is defined as a discrete-continuous hybrid space:
\[
\mathcal{A}=\left\{ \left( k,x_k \right) |x_k\in \varPhi _k\,\,for\,\,all\,\,k\in \left[ K \right] \right\}
\]
where $K$ represents all available phases at an intersection, and $x_k$ is the phase duration associated with the $k$-th phase action. This hybrid action space allows for a flexible adaptation to varying traffic conditions, enabling dynamic control of traffic signals to optimize flow and reduce delays.

The reward function $R$ for the urban traffic signal control is given by:
\[
R = - \sum_{l \in L} q_l
\]
where $q_l$ is the queue length for lane $l$, and $L$ denotes the set of all incoming lanes at the intersection. The reward $R$ represents the negative sum of queue lengths over all incoming lanes, encouraging the reduction of traffic congestion.

\section{Methodology}
\begin{figure*}[htbp]
	\centering
	\includegraphics[width=0.98\textwidth]{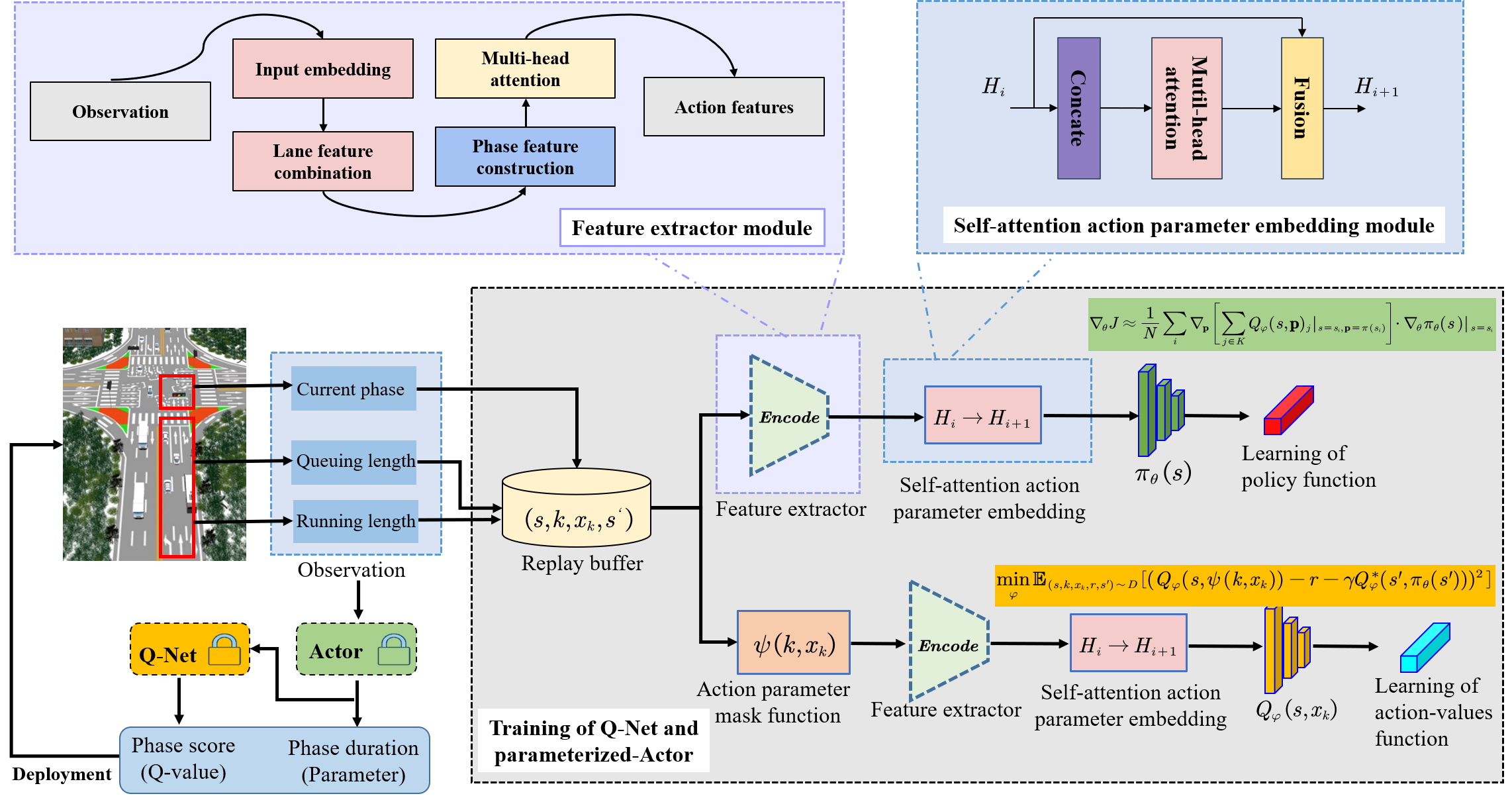}
	\caption{The illustration of our proposed PH-DDPG for ATSC.}
	\label{fig2}
\end{figure*}

This section provides a comprehensive overview of PH-DDPG, which is a novel intelligent traffic control framework leveraging the decoupled parameterized action reinforcement learning with a specific focus on ATSC. We first introduce the action feature representations based on the traffic state and phase. Then we discuss the architecture and implementation of our PH-DDPG framework, as illustrated in Figure \ref{fig2}, which uniquely addresses the challenge of discrete-continuous hybrid action space by decoupling the action parameters for independent optimization. Finally, we design an action parameter mask function, which demonstrates superior effectiveness for the real-world applicability.

\subsection{Action Feature Representations}
Traffic state representation plays an essential role in RL-based TSC models. Motivated by the idea proposed in the AttentionLight \cite{zhang2021attentionlight}, we design a Feature Extractor Module (FEM) to obtain action features based on the observation of queuing vehicles, running vehicles, and the current phase. The process of input embedding is similar to FRAP \cite{zheng2019learning}, where each feature is transformed into a higher-dimensional space. Each phase feature is constructed through the fusion of corresponding lane features, and the correlations between features are learned using a self-attention mechanism. To handle a comprehensive range of action representations, we utilize a Self-attention Action Parameter Embedding Module. The update of the module is given by:

\begin{equation}
	\Gamma_i=Concat\left( H_i,x_k \right),
\end{equation}

\begin{equation}
		H_{i+1} = \alpha \cdot H_i + (1-\alpha ) \cdot \mathrm{Softmax}\left(\frac{Q(\Gamma_i)\cdot K(\Gamma_i)^T}{\sqrt{d_k}}\right) \cdot V(\Gamma_i).
\end{equation}

Here, $H_i$ represents the action features obtained by the FEM, and $x_k$ is the current phase parameter. $\Gamma_i$ is a concatenation of $H_{i}$ and $x_k$, which facilitates the embedding of action features combined with phase parameter information into the network's structure. Both $H_i$ and $\Gamma_i$ are structured as $(\text{batch size}, \text{num\_action}, \text{action features})$, ensuring that all relevant action characteristics are integrated for processing. $H_{i+1}$ is the output of the module, and the coefficient $\alpha$ is a balancing factor that moderates the influence of previous action representations against the newly derived insights from the concatenated representations.

\subsection{Neighboring Nodes Feature Embedding}

To enhance the representation of traffic states, we extend the framework to incorporate data from nearby intersections. This extension enriches the traffic state inputs for each lane at the current intersection by integrating traffic and phase parameters from neighboring nodes. The mathematical formulations for these integrations are given by:
\begin{equation}
    \varGamma_{nb} = \text{Concat}(S_{nb}, x_{nb}, E_{direction}, E_{distance}),
\end{equation}
\begin{equation}
    E_{nb} = \mathrm{Sigmoid}\left(\mathrm{Softmax}\left(\frac{Q(\varGamma_{nb}) \cdot K(\varGamma_{nb})^T}{\sqrt{d_k}}\right) \cdot V(\varGamma_{nb})\right),
\end{equation}
where:
$S_{nb}$ represents the traffic state features from neighboring intersections,
$x_{nb}$ denotes the phase parameters of the neighboring intersections,
$E_{direction}$ is the relative direction embedding, encoding the relative direction between the neighboring node and the current intersection,
$E_{distance}$ is the distance embedding, representing the distance between the neighboring node and the current intersection, and
$\varGamma_{nb}$ is the concatenated vector combining $S_{nb}$, $x_{nb}$, $E_{direction}$, and $E_{distance}$, which integrates traffic state features, phase parameters, and spatial relationships into a unified representation.

The coefficient $\alpha$ balances the influence of prior action representations and new insights derived from the current concatenated inputs. $E_{nb}$ is the newly generated embedding for neighboring nodes, capturing their contextual information. This embedding vector $E_{nb}$ is then integrated into the traffic states for corresponding lanes at the current intersection through:
\begin{equation}
    S = \text{Concat}(S, E_{nb}),
\end{equation}
where $S$ represents the updated traffic state features, now augmented with the neighboring information embedding $E_{nb}$. This enriched representation enables the model to better capture the interdependencies between the current intersection and its neighbors, improving the overall traffic signal control performance.

\subsection{Decoupled Parameterized Action Reinforcement Learning in PH-DDPG}
To address the limitations of traditional reinforcement learning methods in continuous action spaces, we introduce a Q-network $Q_{\varphi}(s, \boldsymbol{x}_k)$ that outputs a vector reward $\hat{\boldsymbol{r}} \in \mathbb{R}^K$. The action parameter vector $\boldsymbol{x}_k = \{ x_{k_1}, x_{k_2}, \ldots, x_{k_K} \}$ encompasses all phase durations of the available discrete phase actions. The expected rewards for each action are associated with the predicted corresponding parameters, expressed as $\hat{\boldsymbol{r}} = Q_{\varphi}(s, \boldsymbol{x}_k)$.

\begin{figure*}[htbp]
    \centering
    \includegraphics[width=0.9\linewidth]{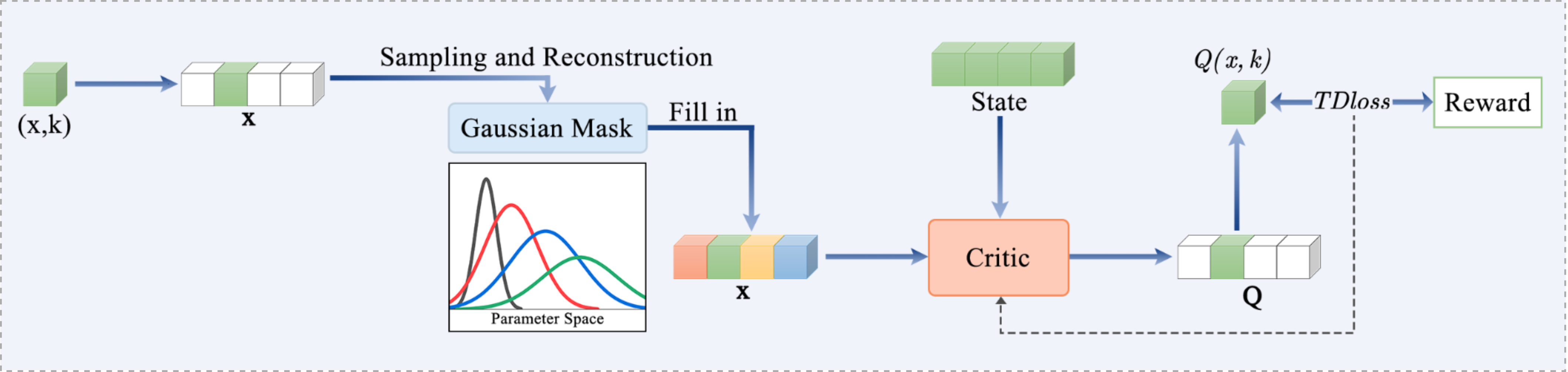}
    \caption{Schematic diagram of the decoupled action space in the PH-DDPG framework. The critic network \(Q_{\varphi}(s, \boldsymbol{x}_k)\) is trained using the reconstructed noisy action parameters, where the action parameter mask function \(\psi(k, x_k)\) selectively masks the parameters of the chosen action \(k\) and introduces Gaussian noise for the parameters of other actions \(j \neq k\).}
    \label{fig:decoupled_action_space}
\end{figure*}

This model departs from traditional scalar reward predictions, facilitating a more granular optimization of action parameters within a discrete-continuous hybrid action space. The network, parameterized by weights $\varphi$, is optimized through a loss function that minimizes the discrepancy between predicted and actual rewards:
\begin{equation}
    \min_{\varphi} \mathbb{E}_{(s, k, x_k, r, s') \sim D} (Q_{\varphi}(s, \psi(k, x_k)) - r - \gamma Q_{\varphi}^*(s', \pi_{\theta}(s'))) ^2
    \label{eq5}
\end{equation}

where $Q_{\varphi}^*(s', \pi_{\theta}(s'))$ denotes the maximum expected reward for the next state $s'$. To further illustrate the decoupled action space and the masking process during the critic network training, we refer to Figure~\ref{fig:decoupled_action_space}. The figure provides a schematic representation of how the action parameter vector \(\boldsymbol{x}_k\) is decoupled and processed within the critic network. Specifically, during the training of the critic network, the action parameter mask function \(\psi(k, x_k)\) is employed to reconstruct the noisy action parameters. This is achieved by selectively masking the action parameters associated with the chosen action \(k\) while introducing Gaussian noise \(\epsilon\) for the parameters corresponding to other actions \(j \neq k\). The noise is sampled from a normal distribution \(\mathcal{N}(\mu_{\varPhi_k}, \sigma_{\varPhi_k}^2)\), where \(\mu_{\varPhi_k}\) and \(\sigma_{\varPhi_k}^2\) are derived from the distribution \(\varPhi_k \sim D_{N}\).
The action parameter mask function $\psi(k, x_k)$ is defined as:
\begin{equation}
    \boldsymbol{x}_j = \psi(k, x_k) = 
    \begin{cases} 
        x_k, & \text{if } j = k, \\
        \epsilon \sim \mathcal{N}(\mu_{\varPhi_k}, \sigma_{\varPhi_k}^2),\varPhi_k \sim D_{N} , & \text{if } j \neq k,
    \end{cases}
    \label{eq6}
\end{equation}
where $\epsilon$ is Gaussian noise sampled from a normal distribution parameterized by the mean $\mu_{\varPhi_k}$ and variance $\sigma_{\varPhi_k}^2$. These parameters are derived from the distribution $\varPhi_k \sim D_{N}$, where $D_{N}$ represents the data distribution over a batch of size $N$. Specifically, $N$ denotes the batch size used during training, and $\varPhi_k$ is sampled from the empirical distribution of the batch data. This ensures that the Gaussian noise introduced for exploration is calibrated to the statistical properties of the training data, enhancing the stability and effectiveness of the exploration process.
% where $\epsilon$ is Gaussian noise sampled from a distribution with mean $\mu_{\varPhi_k}$ and variance $\sigma_{\varPhi_k}^2$, which are derived from the distribution $\varPhi_k \sim D_{N}$. Here, $\mathcal{N}(\mu_{\varPhi_k}, \sigma_{\varPhi_k}^2)$ represents the normal distribution parameterized by the mean $\mu_{\varPhi_k}$ and variance $\sigma_{\varPhi_k}^2$ of $\varPhi_k$.

% \madfddff{df1}
The parameterized actor, denoted as $\pi_{\theta}(s)$, is designed to deduce the optimal configuration of action parameters $\boldsymbol{x}_k^* = \{x_{k_1}^*,x_{k_2}^*,\cdots ,x_{k_K}^*\}= \pi_{\theta}(s)$. We rigorously prove that minimizing the negative sum of accumulated estimated rewards $- \sum_{k\in K} \hat{\boldsymbol{r}}_k$ identifies an vector of action parameters $\boldsymbol{x}_k^*$, which maximizes the expected reward obtained for each action executed in state $s$.

We use a deterministic policy gradient approach, wherein the policy is directly shaped by the gradients obtained from the Q-network relative to the action parameters. The optimization of the parameters in actor network during back-propagation is encapsulated by the following policy gradient formulation:
\begin{equation}
    \begin{aligned}
    \nabla_{\theta} J \approx \frac{1}{N} \sum_i \nabla_{\boldsymbol{x}_k} \left[\sum_{j\in  K} Q_{\varphi}(s, \boldsymbol{x}_k)_j|_{s=s_i, \boldsymbol{x}_k=\pi(s_i)}\right] \\
    \cdot \nabla_{\theta} \pi_{\theta}(s)|_{s=s_i}.
    \end{aligned}
    \label{eq7}
\end{equation}
This equation defines the gradient of the objective function $J$ with respect to the actor weights $\theta$. 

To formally establish the validity of this approach,  we need prove that minimizing the negative sum of accumulated estimated rewards $- \sum_{k\in K} \hat{\boldsymbol{r}}_k$ identifies a vector of action parameters, which maximizes the expected reward obtained for each action. First, let $x$ denote the action parameters, defined within a finite, simply connected real space, and let the action-value function $Q(s,a)$ be continuously differentiable with respect to $x$:

\textbf{Assumptions:}

\begin{enumerate}
	
	\item $\forall i \in K, \exists x_i^*\in \varPhi_i: x_i^* = \arg \max_{x_i} \mathbb{E}_{(s,r) \sim E}(r|s,a_i,x_i)$. 
	
	\item $\forall i \in K, \forall x_j \in \varPhi_j, j \in K \setminus \{i\}: \mathbb{E}_{(s,r) \sim E}(r|s,a_i,x_i)  = Q_{\varphi}(s,\boldsymbol{x}|_{\boldsymbol{x}_i=x_i , \boldsymbol{x}_j=x_j})_i$.
\end{enumerate}

\textbf{Proof:}
\begin{enumerate}
	\item[1] From assumption 2, since the value of $\hat{\boldsymbol{r}}_i$ depends solely on the corresponding action parameter $x_i$ and is independent of other action parameters $x_j, for j \neq i$, we always have: 
	\[
	\frac{\partial \hat{\boldsymbol{r}}_i}{\partial x_j} = 0, \quad \forall j \neq i.
	\]
	
	Then, from assumption 2, we derive: 
	\[
	\forall i \in K :\ \  \mathbb{E}_{(s,r) \sim E}(r|s,a_i,x_i) = Q_{\varphi}(s,\boldsymbol{x}|_{\boldsymbol{x}_i=x_i})_i
	\]
	
	\item[2] Let us consider the possibility of an alternative parameter array $\boldsymbol{x}' \neq \boldsymbol{x}^*$ where $\sum_{i\in K} \hat{\boldsymbol{r}}_i|\boldsymbol{x}' > \sum_{i\in K}  \hat{\boldsymbol{r}}_i|\boldsymbol{x}^*$. we have that:
	\[
	\exists i \in K: \hat{\boldsymbol{r}}_i|\boldsymbol{x}' > \hat{\boldsymbol{r}}_i|\boldsymbol{x}^*
	\]
	Let $x_i'$ be denoted by $(\boldsymbol{x}')_i$, and based on Proof 1, we have: 
	
	\[
	\exists i \in K:\  Q_{\varphi}(s,\boldsymbol{x}|_{\boldsymbol{x}_i=x_i'})_i > Q_{\varphi}(s,\boldsymbol{x}|_{\boldsymbol{x}_i=x_i^*})_i
	\]

	This is equivalent to:
	\[
    \begin{aligned}
        &\exists i \in K, \exists x_i' \in \varPhi_i: \\
        &\quad \mathbb{E}_{(s,r) \sim E}(r|s,a_i,x_i') > \mathbb{E}_{(s,r) \sim E}(r|s,a_i,x_i^*)
    \end{aligned}
	\]

	contradicting assumption 1. It is concluded that there does not exist any parameter array $\boldsymbol{x}'$ that achieves a higher value of $\sum_{i\in K} \hat{\boldsymbol{r}}_i$ than $\boldsymbol{x}^*$, affirming the superiority of $\boldsymbol{x}^*$ in maximizing the accumulated estimated rewards.
	
\end{enumerate}

\subsection{The Implementation of PH-DDPG}
The implementation of PH-DDPG can be divided into dual modes as delineated in Algorithm \ref{al1}. This approach distinguishes between two methodologies based on their differing strategies for initializing and updating the replay buffer $D$.

\textbf{Offline mode:} In this configuration, the replay buffer $D$ is populated with a comprehensive dataset derived from historical data, reflecting a broad spectrum of environmental states. Transitions $(s,k,x_k,r,s')$, generated via established decision protocols, populate $D$ for foundational training.

\textbf{Online mode:} This mode initiates with a limited dataset in $D$, which is then incrementally enhanced with data from real-time interactions. This process involves sampling environmental states, optimizing action parameters $\boldsymbol{x}^*$ through the actor network $\pi_{\theta}(s)$, and calculating corresponding expected rewards. Subsequently, new transitions $(s,k,x^*_k,\boldsymbol{r}_k,s')$ are assimilated into $D$.

Experimental analyses are presented to comparatively assess their efficacy under distinct mode.  

\begin{algorithm}[htbp]
\small
\caption{\small{PH-DDPG Online/Offline Algorithm}}\label{al1}
\begin{algorithmic}[1]
\Require Discrete-continuous hybrid action set $\mathcal{A}$ indexed by $(k,x_k)$, maximum episodes $M$, maximum time steps $T$, interaction number $H$ per episode, update frequency $d$, target network update rate $\tau$
\Ensure Well-trained Q-network $Q_{\varphi}(s,\boldsymbol{x}_k)$ and actor $\pi_{\theta}(s)$

\State Initialize Q-network $Q_{\varphi}(s,\boldsymbol{x}_k)$ and actor $\pi_{\theta}(s)$ with random parameters $\varphi$ and $\theta$
\State Initialize target networks $Q_{\varphi'}$ and $\pi_{\theta'}$ with $\varphi' \leftarrow \varphi$, $\theta' \leftarrow \theta$
\State Initialize replay buffer $D$ with transitions $(s,k,x_k,r,s')$ from predefined decision scheme

\For{episode = 1 to $M$}
    \For{step = 1 to $T$}
        \State Sample batch of $N$ experiences $(s,k,x_k,r,s')$ from $D$
        \State Compute $\boldsymbol{x}_k' \leftarrow \pi_{\theta'}(s')$
        \State Construct action parameter mask $\psi(k,x_k)$ using \eqref{eq6}
        \State Update Q-network $Q_{\varphi}$ using \eqref{eq5}
        
        \If{step mod $d$ == 0}
            \State Compute $\hat{\mathbf{r}} = Q_{\varphi}(s,\boldsymbol{x}_k,\pi_{\theta}(s))$
            \State Update actor $\pi_{\theta}$ using \eqref{eq7}
            \State Update target networks:
            \State $\quad \varphi' \leftarrow \tau\varphi + (1-\tau)\varphi'$
            \State $\quad \theta' \leftarrow \tau\theta + (1-\tau)\theta'$
        \EndIf
    \EndFor
    
    \If{Online mode == True}
        \For{interaction = 1 to $H$}
            \State Sample state $s$ from environment
            \State Compute $\boldsymbol{x}^* = \pi_{\theta}(s)$ and $\boldsymbol{r} = Q_{\varphi}(s,\boldsymbol{x}^*)$
            \State Select action $k \leftarrow \arg\max_{k\in[K]} \boldsymbol{r}_k$
            \State Execute action $(k,x^*_k)$, observe reward $\boldsymbol{r}_k$ and next state $s'$
            \State Store transition $(s,k,x^*_k,\boldsymbol{r}_k,s')$ in $D$
        \EndFor
    \EndIf
\EndFor
\end{algorithmic}
\end{algorithm}

\section{Performance Evaluation}
In this section, we conduct extensive experiments to demonstrate how our proposed model performs compared to other state-of-the-art methods. The experimental datasets strictly follow the real-world traffic characteristics. In addition to using the most frequently measured metric, we have also developed two novel metrics that better align with real-world application evaluations. 

\subsection{Real-world Datasets}

\begin{figure*}[t!]
  \centering
  \begin{tabular}{ccc}
  \includegraphics[width=0.178\textwidth]{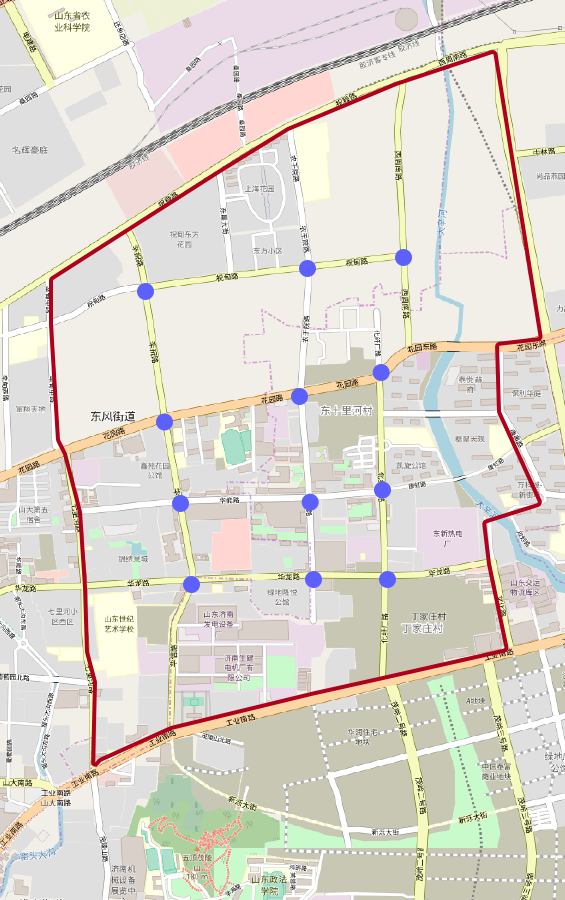}&
  \includegraphics[width=0.38\textwidth]{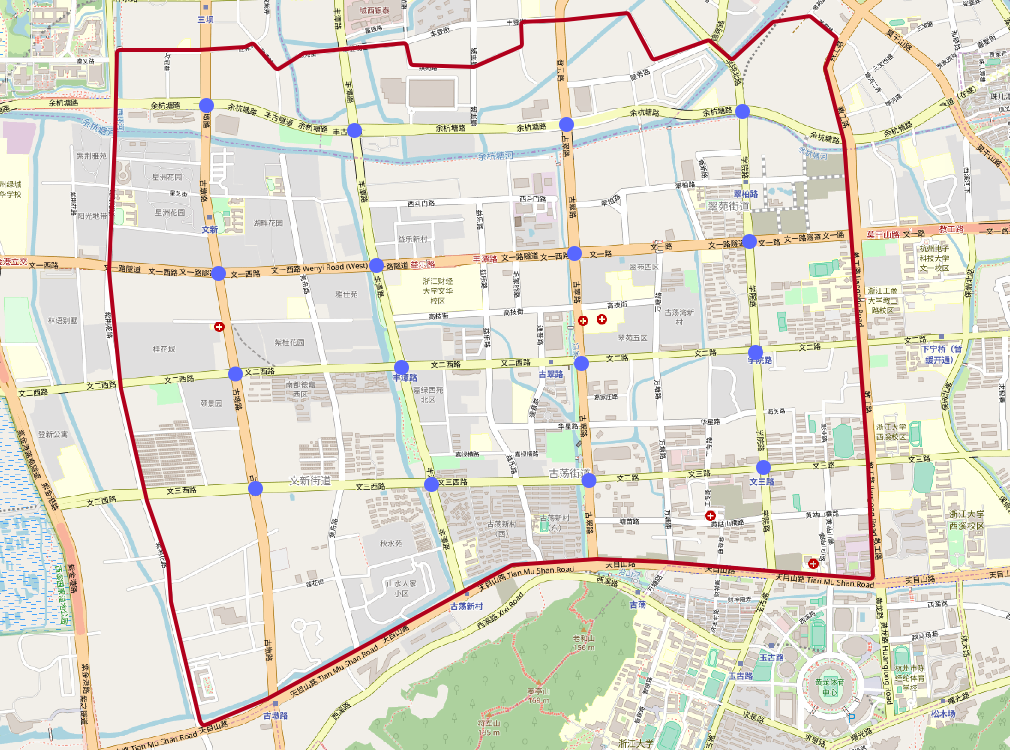}&
  \includegraphics[width=0.20\textwidth]{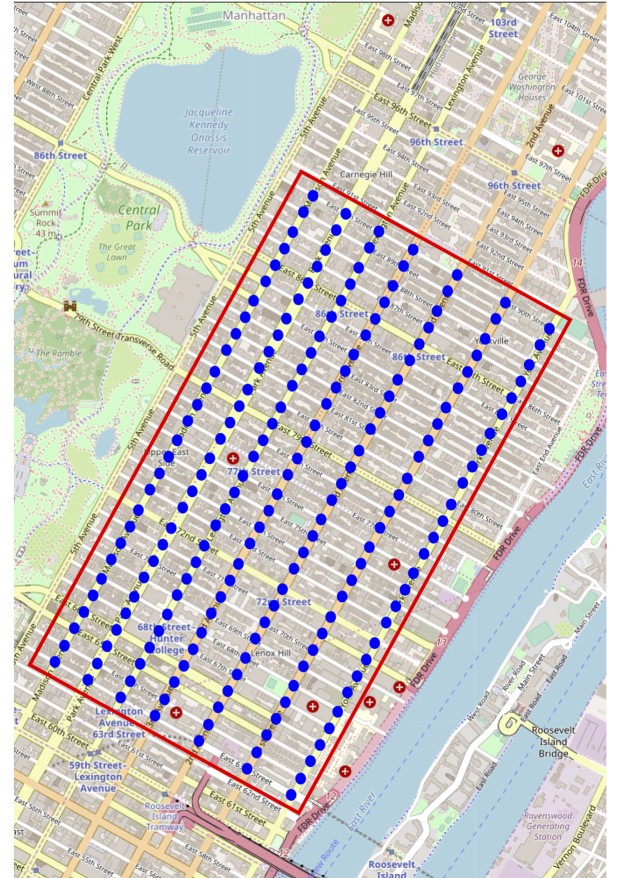}\\
  \begin{tabular}[c]{@{}c@{}}(a) Dongfeng Sub-district,\\ Jinan, China\end{tabular} &
  \begin{tabular}[c]{@{}c@{}}(b) Gudang Sub-district,\\ Hangzhou, China\end{tabular} &
  \begin{tabular}[c]{@{}c@{}}(c) Upper East Side, Manhattan,\\New York, USA\end{tabular}\\
  \end{tabular}
  \caption{Study area road networks with controlled traffic signals (blue dots). (a) 12-intersection network with bidirectional traffic in Jinan; (b) 16-intersection system supporting uni- and bi-directional flows in Hangzhou; (c) 196-signal unidirectional network in Manhattan. Red polygons denote modeled areas.}
  \label{fig:real_road_net}
  \vspace{-3mm}
\end{figure*}

We utilize real-world traffic data from three cities: Jinan, Hangzhou, and New York. The road networks are obtained from OpenStreetMap, as illustrated in Figure~\ref{fig:real_road_net}. The Jinan dataset covers 12 intersections in Dongfeng Sub-district, while Hangzhou includes 16 intersections in Gudang Sub-district, both collected through roadside surveillance systems. The New York dataset comprises 196 intersections in Manhattan's Upper East Side, with traffic patterns derived from processed taxi trip records. All datasets contain detailed information about lanes, roads, intersections, and signal phases, with traffic flow represented by $(t, u)$, where $t$ indicates the entry time and $u$ represents the pre-determined route.

\subsection{Experimental Settings}

In the experiment, we use the real-world traffic data from three cities: Jinan, Hangzhou and New York. Each dataset consists of lanes, roads, intersections, traffic flow and signal phases. The traffic flow data describes the movement of vehicles via $(t, u)$, where $t$ marks the entry time and $u$ its pre-determined route. For a fair comparison, all the models are executed on a workstation with the following specifications: 126GB memory, Intel Xeon(R) W-2255 CPU @ 3.70GHz × 20, and an NVIDIA GeForce RTX 3090 GPU card. We optimize all the models using the Adam optimizer for a maximum of 100 episodes. Learning rates are adjusted accordingly, and batch sizes are optimised for computational efficiency, $\gamma = 0.8$. For online reinforcement learning, performance results are averaged over the last 10 episodes. For offline training of PH-DDPG and its variants, the replay buffer is initialized with data from 30 episodes and trained for 40 episodes. Each phase ends with a three-second yellow light followed by a two-second red light.

\subsection{Comparative Methods}

We compare our model with the following two categories of methods: conventional transportation methods and reinforcement learning (RL) methods.

% \paragraph{The traditional TSC methods}

\textbf{Traditional methods}: FixedTime \cite{koonce2008traffic} uses a pre-determined plan for cycle length and phase duration, which is widely used when the traffic flow is steady. MaxPressure \cite{varaiya2013max} is a method for network-level traffic signal control, which uses a greedy algorithm to choose the phase that maximizes pressure.

\textbf{RL methods}: FRAP \cite{zheng2019learning} is another state-of-the-art RL-based method with a modified network to capture the phase competition relation between different traffic movements. MP-DQN \cite{bester2019mpdqn} addresses parameterised action-space challenges through multi-pass separation of action-parameters, while in this work we specifically utilize phase timing as the action parameter. CoLight \cite{wei2019colight} employs a graph attention network to foster cooperation between intersections. Advanced-CoLight \cite{zhang2022expression} adds the advanced traffic states to the observation of CoLight. AttentionLight \cite{zhang2021attentionlight} uses self-attention mechanisms to refine phase feature construction. DynamicLight \cite{zhang2022dynamiclight} combines MQL for phase selection with a deep RL network for timing adjustments. DataLight \cite{zhang2023data} uses a CQL-based approach for offline training with cyclical, expert, and random datasets.

\begin{table*}[h]
	\caption{\textbf{Comparison of performance on the Jinan dataset}. The higher DAR values indicate larger vehicle throughput (the higher, the better). The lower ATT or DATT values indicate less waiting time (the lower, the better). \textbf{Best} results are bolded.}
	\label{Jinan_data}
	\centering
	\resizebox{\linewidth}{!}{
		\begin{tabular}{llccccccccc}
			\toprule
			& \multirow{4}{*}[3mm]{Method} & \multicolumn{3}{c}{Jinan-1} & \multicolumn{3}{c}{Jinan-2} & \multicolumn{3}{c}{Jinan-3}\\
			\cmidrule(lr){3-5} \cmidrule(lr){6-8} \cmidrule(lr){9-11}
			& & ATT & DATT & DAR & ATT & DATT & DAR & ATT & DATT & DAR \\
			\midrule
			\multirow{2}{*}{Traditional} 
			& FixedTime & 499.9 & 463.3 & 0.7919 & 412.6 & 414.3 & 0.9099 & 445.7 & 438.4 & 0.8407 \\
			& MaxPressure & 319.9 & 329.4 & 0.9018 & 293.2 & 298.9 & 0.9486 & 291.0 & 300.5 & 0.9114\\
			\cmidrule{1-11}
			\multirow{6}{*}{Online RL}
                & MP-DQN & 1014.3 & 591.4 & 0.3720 & 987.6 & 587.2 & 0.4872 & 1006.5 & 601.0 & 0.3981 \\ 
			& Colight & 311.8 & 320.8 & 0.9088 & 295.9 & 301.5 & 0.9464 & 292.6 & 302.0 & 0.9086\\
			& QL-FRAP & 302.2 & 311.2 & 0.9123 & 287.3 & 292.9 & 0.9498 & 283.8 & 293.0 & 0.9137\\ 
			& AttentionLight & 302.2 & 311.4 & 0.9109 & 286.9 & 292.6 & 0.9509 & 282.9 & 291.9 & 0.9119\\
			& Advanced-CoLight & 290.8 & 299.2 & 0.9132 & 280.5 & 285.8 & 0.9505 & 275.6 & 284.4 & 0.9155\\
			& DynamicLight & 287.6 & 296.0 & 0.9167 & 277.3 & 282.5 & 0.9507 & 273.6 & 282.3 & 0.9164 \\
			& PH-DDPG & \textbf{284.8} & \textbf{292.9} & \textbf{0.9171} & \textbf{273.7} & \textbf{275.8} & \textbf{0.9525} & \textbf{268.4} & \textbf{276.7} & \textbf{0.9183}\\
			& PH-DDPG-NB & \textbf{283.6} & \textbf{291.9} & \textbf{0.9186} & \textbf{271.8} & \textbf{274.3} & \textbf{0.9609} & \textbf{267.5} & \textbf{275.9} & \textbf{0.9191}\\
			\cmidrule{1-11}
			\multirow{3}{*}{Offline RL}  
			& DataLight (random) & 290.8 & 299.4 & 0.9151 & 279.6 & 284.7 & 0.9507 & 277.4 & 286.2 & 0.9157 \\
			& DataLight (expert) & 288.6 & 296.9 & \textbf{0.9155} & 277.3 & 282.6 & 0.9521 & 273.7 & 282.1 & 0.9164 \\
			& PH-DDPG (random) & \textbf{286.2} & \textbf{294.6} & 0.9151 & \textbf{275.7} & \textbf{280.9} & \textbf{0.9521} & \textbf{270.8} & \textbf{279.3} & \textbf{0.9180}\\
			\bottomrule
	\end{tabular}}
\end{table*}

\begin{table*}[h]
	\caption{\textbf{Comparison of performance on the Hangzhou dataset}. The higher DAR values indicate larger vehicle throughput (the higher, the better). The lower ATT or DATT values indicate less waiting time (the lower, the better). \textbf{Best} results are bolded.}
	\label{HangZhou_data}
	\centering
    \resizebox{0.8\linewidth}{!}{
	\begin{tabular}{llcccccc}
		\toprule
		%   \multicolumn{2}{c}{Part}            qa        \\
		& \multirow{4}{*}[3mm]{Method} & \multicolumn{3}{c}{Hangzhou-1} & \multicolumn{3}{c}{Hangzhou-2} \\
		\cmidrule(lr){3-5} \cmidrule(lr){6-8} 
		& & ATT & DATT & DAR & ATT & DATT & DAR \\
		\midrule
		\multirow{2}{*}{Traditional} 
		& FixedTime & 568.8 & 541.6 & 0.8176 & 559.5 &452.7& 0.5453 \\
		& MaxPressure & 334.2 & 344.7 & 0.9165 & 442.1 & 404.8 & 0.6268 \\
		\cmidrule{1-8}
		\multirow{6}{*}{Online RL} 
            & MP-DQN & 1241.2 & 673.5 & 0.4023 & 1311.7 & 696.4 & 0.3011 \\
		& Colight & 338.4 & 348.0 & 0.9135 & 425.5 & 391.9 & 0.6492 \\
		& QL-FRAP & 328.1 & 337.9 & 0.9165 & 413.0 & 375.4 & 0.6569\\
		& AttentionLight & 328.3 & 338.1 & 0.9155 & 414.1 & 374.7 & 0.6531\\
		& Advanced-CoLight & 316.7 & 326.5& 0.9185 & 408.3 & 389.2 & 0.6623 \\
		& DynamicLight  & 316.2 & 325.4 & 0.9208 & 406.1 & 384.8 & 0.6672\\
		& PH-DDPG & \textbf{308.1} & \textbf{317.1} & \textbf{0.9226} & \textbf{398.2} & \textbf{374.3} & \textbf{0.6782} \\
		& PH-DDPG-NB & \textbf{307.9} & \textbf{317.2} & \textbf{0.9235} & \textbf{399.5} & \textbf{369.2} & \textbf{0.6681}\\
		\cmidrule{1-8}
		\multirow{3}{*}{Offline RL} 
		& DataLight (random) & 318.6 & 327.9 & 0.9192 & 402.1 & 372.8 & 0.6652 \\
		& DataLight (expert) & 314.2 & 323.5 & 0.9205 & 398.2 & 381.7 & \textbf{0.6755}\\
		& PH-DDPG (random) & \textbf{309.3} & \textbf{318.3} & \textbf{0.9219} & \textbf{397.5} & \textbf{377.9} & 0.6729 \\
		\bottomrule
	\end{tabular}}
\end{table*}

\begin{table*}[h]
	\caption{\textbf{Comparison of performance on the New York dataset}. The higher DAR values indicate larger vehicle throughput (the higher, the better). The lower ATT or DATT values indicate less waiting time (the lower, the better). \textbf{Best} results are bolded.}
	\label{New York_data}
	\centering
    \resizebox{0.8\linewidth}{!}{
	\begin{tabular}{llcccccc}
		\toprule
		%   \multicolumn{2}{c}{Part}                   \\
		& \multirow{4}{*}[3mm]{Method} & \multicolumn{3}{c}{New York-1} & \multicolumn{3}{c}{New York-2} \\
		\cmidrule(lr){3-5} \cmidrule(lr){6-8} 
		& & ATT & DATT & DAR & ATT & DATT & DAR \\
		\midrule
		\multirow{2}{*}{Traditional} 
		& FixedTime & 1466.5 & 1047.6 & 0.1850 & 1581.7 &1041.1& 0.1139 \\
		& MaxPressure & 1229.3 & 917.4 & 0.3235 & 1440.7 & 865.9 & 0.1797 \\
		\cmidrule{1-8}
		\multirow{6}{*}{Online RL} 
            & MP-DQN & 1786.1 & 1147.6 & 0.1102 & -  & - & - \\
		& Colight & 1096.3 & 697.6 & 0.3657 & 1370.0& 755.2& 0.2136\\
		& QL-FRAP & 1105.0 & 778.7 & 0.3851 & 1511.4 & 794.1 & 0.1314\\
		& AttentionLight & 1194.5 & 677.0 & 0.2931 & 1491.5 & 778.4 & 0.1438\\
		& Advanced-CoLight & 1058.5 & \textbf{638.3} & 0.3942 & 1304.1 & 721.8 & 0.2489 \\
		& DynamicLight  & 1145.1 & 791.7 & 0.3441 & 1397.6 & \textbf{719.5} & 0.1944\\
		& PH-DDPG & 1070.1 & 694.9 & \textbf{0.3999}  & 1318.0 & 758.9 & \textbf{0.2511} \\
		& PH-DDPG-NB & \textbf{976.4} & 730.1 & \textbf{0.4918}  & \textbf{1233.0} & 789.2 & \textbf{0.3075} \\
		\cmidrule{1-8}
		\multirow{1}{*}{Offline RL}
		& PH-DDPG (random) & \textbf{1092.8} & \textbf{704.8} & \textbf{0.3728} & \textbf{1337.1} & \textbf{641.7} & 0.2237 \\
		\bottomrule
	\end{tabular}}
\end{table*}

\begin{figure*}[htbp]
	\centering
	\includegraphics[width=0.9\textwidth]{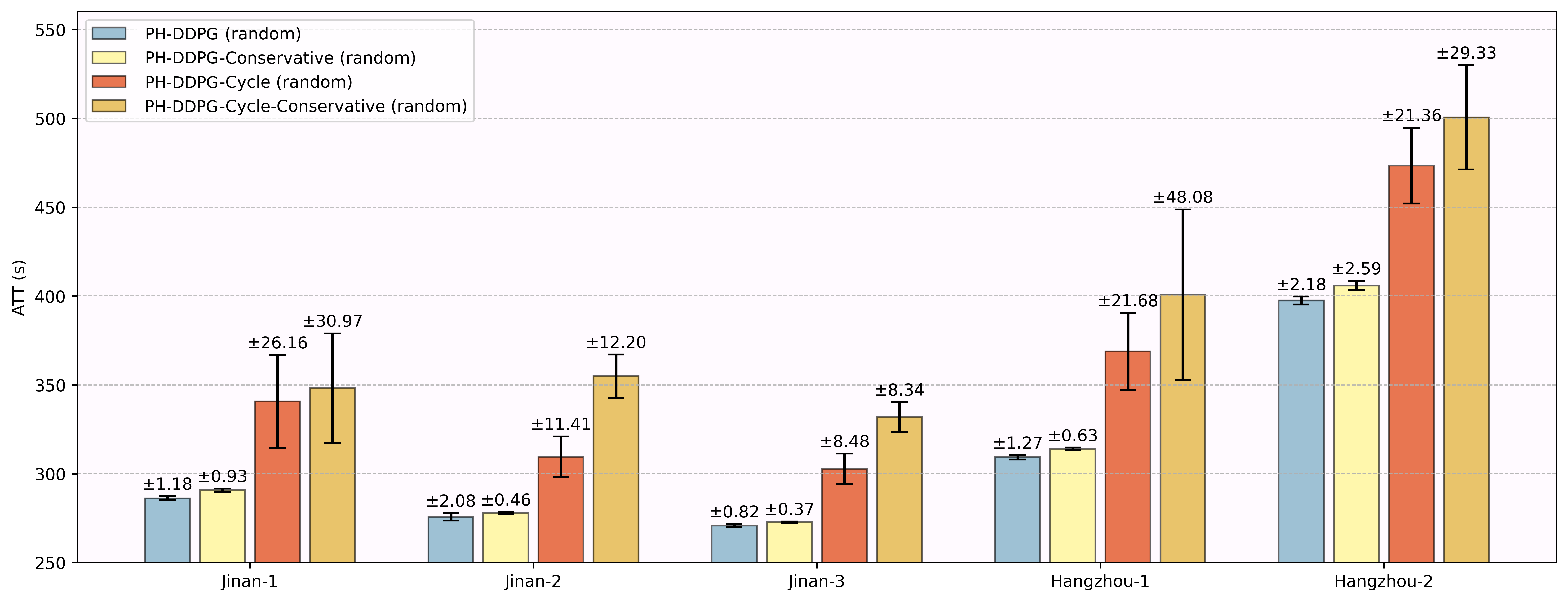}
 
	\caption{ATT with error bars representing model variance under Jinan and Hangzhou datasets.}
	\label{fig4}
\end{figure*}

\subsection{Evaluation Metrics}

Following existing studies as mentioned above, we use the average travel time (\textbf{ATT}) to evaluate the performance of different models. It calculates the average travel time of all vehicles in the system, which is the most frequently used measure in transportation. However, we have observed that the traffic backlogs are prematurely cleared at the end of simulation when calculating the ATT, which induces underestimation of results and cannot evaluate model performance fairly. Therefore, we design the destination-arrival average travel time (\textbf{DATT}) and the destination-arrival rate (\textbf{DAR}), that can conduct more accurate assessments and provide a clearer insight into the efficacy of traffic signal control strategies.

\textbf{DATT}: It measures the average travel time for vehicles that have reached their destinations, and directly indicates the effectiveness of TSC strategies in reducing travel times for commuters.

\textbf{DAR}: It calculates the proportion of vehicles arriving at their destinations relative to the total number of vehicles over a specific period. It provides insights into the effectiveness of TSC methods in managing traffic flow during specific periods. In all experiments in this paper, we calculate the DAR in an hour, which allows for a consistent comparison across different datasets.

\subsection{Performance Comparison}
\label{compar}

To fully evaluate the effectiveness of our method, we have generated \textbf{PH-DDPG-NB}, \textbf{PH-DDPG-Cycle}, \textbf{PH-DDPG-Conservative}, and \textbf{PH-DDPG-Conservative-Cycle} by our \textbf{PH-DDPG} algorithm template. These methods are designed to enhance signal phase durations according to real-time traffic conditions, with variations tailored to specific operational needs and scenarios.

\textbf{PH-DDPG-NB}: Integrates information from nearby intersections into the traffic state input for each lane at the current intersection, enhancing the model's responsiveness to dynamic traffic conditions.

\textbf{PH-DDPG-Cycle}: Derived from the original framework, it emphasizes cyclical phase transitions that either sustain the present phase or transition to a predetermined subsequent phase, closely simulating real-world scenarios. The core idea involves determining the optimal action $i$ and its best parameter $x$ given the current action $i_{\text{current}}$ and the next action $i_{\text{next}}$:

    \begin{equation}
        (i, x) = 
        \begin{cases} 
            (i_{\text{current}}, x_{i_{\text{current}}}), & \text{if } Q_{\varphi}(s, \boldsymbol{x}_k)_{i_{\text{current}}} \geq Q_{\varphi}(s, \boldsymbol{x}_k)_{i_{\text{next}}}, \\
            (i_{\text{next}}, x_{i_{\text{next}}}), & \text{if } Q_{\varphi}(s, \boldsymbol{x}_k)_{i_{\text{current}}} < Q_{\varphi}(s, \boldsymbol{x}_k)_{i_{\text{next}}},
        \end{cases}
        \label{eq_cycle}
    \end{equation}
    
    where $i_{\text{current}}$ represents the current action, and $i_{\text{next}}$ represents the subsequent action. $Q_{\varphi}(s, \boldsymbol{x}_k)$ denotes the Q-value function, evaluated at state $s$ with action parameter $\boldsymbol{x}_k$. 

\textbf{PH-DDPG-Conservative}: Aligning with the foundational framework, it constrains temporal adjustments within a 5-second margin, a deliberate design choice tailored for preliminary real-world applications to mitigate substantial fluctuations in signal temporalities.

\textbf{PH-DDPG-Conservative-Cycle}: This version amalgamates the conservative time adjustment methodology and phase selection techniques of the Conservative and Cycle iterations, culminating in a comprehensive model tailored for the seamless adoption of PH-DDPG in authentic traffic settings.

\subsubsection{Overall Performance}

Table \ref{Jinan_data} and Table \ref{HangZhou_data} demonstrate the overall performance under multiple real-world datasets concerning the ATT, DATT, and DAR. PH-DDPG achieves a new state-of-the-art performance. While MP-DQN similarly addresses parameterised actions in a parallel framework, it demonstrates limited effectiveness when handling phase timing as action parameters, particularly exhibiting persistent convergence failures in the New York dataset (Table \ref{New York_data}). This suggests the necessity for further adaption efforts of MP-DQN in traffic control applications. In offline mode, PH-DDPG exhibits superior robustness, outperforming DataLight even when trained on random datasets. This feature underscores the flexibility and efficacy of PH-DDPG, diminishing the need for expert datasets and expediting the deployment of traffic management solutions. Compared to DataLight (expert), PH-DDPG reduces ATT by 1.56\% and 0.18\%, and reduces DATT by 1.61\% and 1.00\% in Hangzhou-1 and Hangzhou-2, respectively. These findings underscore the inherent advantage of PH-DDPG in obviating the need for expert data, thereby streamlining the deployment process of algorithms. Additionally, the performance on the New York dataset further validates the effectiveness of PH-DDPG, particularly in complex urban settings. As shown in Table \ref{New York_data}, PH-DDPG significantly outperforms Advanced-CoLight in both New York-1 and New York-2, achieving notable reductions in ATT by 7.76\% and 5.45\%, and enhancements in DAR by 9.76\% and 5.86\%. This performance is primarily due to PH-DDPG's ability to utilize state information from a single intersection and effectively incorporate information from neighboring nodes, addressing the limitations faced in densely interconnected urban areas.

\begin{figure*}[htb]
	\centering
	\includegraphics[width=0.95\textwidth]{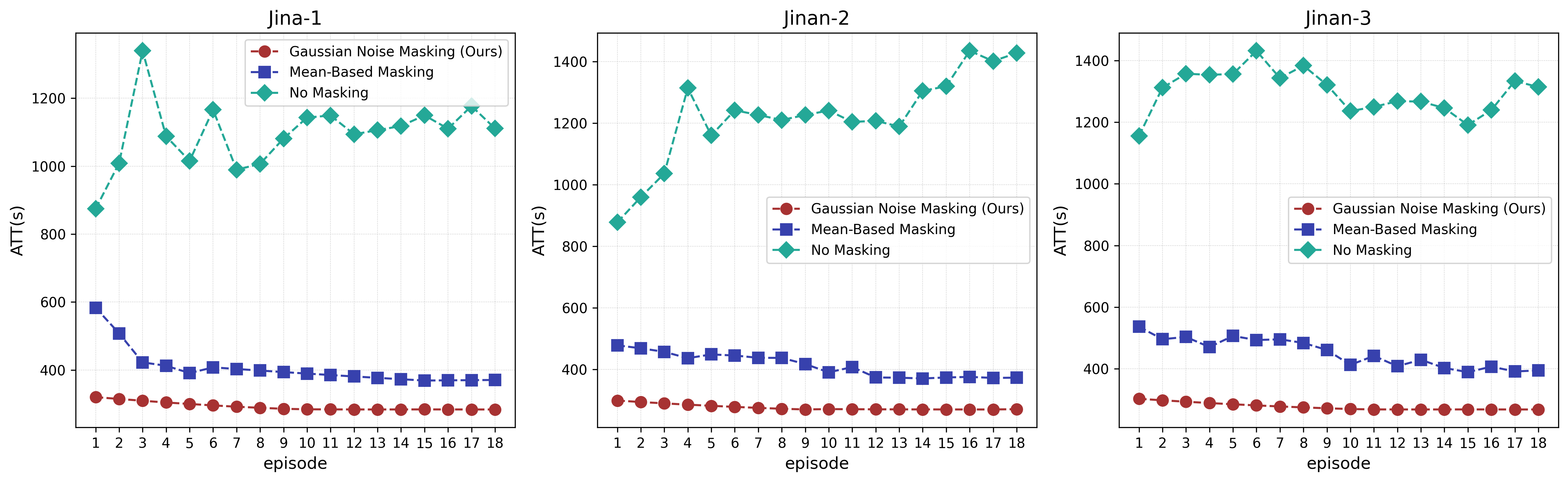}
	\caption{Evolution of averaged travel time (ATT) across three Jinan datasets during PH-DDPG offline training with random exploration dataset. Model checkpoints from each training episode were evaluated in real traffic scenarios, comparing three masking strategies: (1) Gaussian noise masking (red), (2) constant mean-value masking (blue), and (3) no masking (green).}
	\label{fig5}
\end{figure*}

\begin{table*}[htb]
	\caption{\textbf{Sensitivity Analysis of Mini-batch Size $ N $ on Jinan1/2/3}. Values represent mean $\pm$ standard deviation of ATT (seconds) across 4 experimental runs. The episode column indicates the mean episode at which the optimal ATT was achieved across the three datasets.}
	\label{tab:N_sensitivity}
    \centering
	\resizebox{\linewidth}{!}{
	\begin{tabular}{lcccc}
		\toprule
		$ N $ (Mini-batch) & \multicolumn{1}{c}{Jinan-1 ATT (s)} & \multicolumn{1}{c}{Jinan-2 ATT (s)} & \multicolumn{1}{c}{Jinan-3 ATT (s)} & \multicolumn{1}{c}{Optimal ATT Episode (mean)} \\
		\midrule
		20  & $ 421.2 \pm 58.2 $ & $ 296.3 \pm 5.9 $  & $ 289.8 \pm 4.2 $  & 74 \\
		60  & $ 289.8 \pm 2.3 $  & $ 279.4 \pm 3.9 $  & $ 273.2 \pm 1.3 $  & 18 \\
		80 (Ours) & $ 286.2 \pm 1.2 $  & $ 276.2 \pm 2.1 $  & $ 270.8 \pm 0.8 $  & 26 \\
		120 & $ 285.9 \pm 0.9 $  & $ 275.8 \pm 1.7 $  & $ 270.1 \pm 0.6 $  & 48 \\
		250 & $ 285.1 \pm 0.6 $  & $ 276.0 \pm 1.0 $  & $ 269.6 \pm 0.3 $  & 72 \\
		\bottomrule
	\end{tabular}}
\end{table*}

\subsubsection{Evaluation of PH-DDPG Variants in Offline Mode}

As illustrated in Figure \ref{fig4}, the performance of the PH-DDPG models across various datasets reveals distinct strengths in their effectiveness and reliability. The PH-DDPG model, facilitating optimal phase selection, consistently outperforms by showcasing the lowest ATT and minimal variability across all datasets. Similarly, the PH-DDPG-Conservative model, although exhibiting slightly higher ATT values, maintains very low variances, highlighting its reliability in initial real-world deployments. In contrast, the PH-DDPG-Cycle and PH-DDPG-Conservative-Cycle models, designed to mimic real-world cyclic phase transitions, show higher ATT values and larger variances. This suggests less stability and efficiency, indicating a need for further optimization to enhance their performance in complex traffic environments. In summary, the dynamic selection of optimal phases emerges as the pivotal element in enhancing traffic management efficiency. While conservative strategies ensure consistent performance with minimal interference, cyclic models necessitate further fine-tuning to attain similar levels of stability and efficacy.

\subsection{Ablation Study}

Figure \ref{fig5} illustrates the variations in averaged travel time (ATT) during the training of PH-DDPG across three Jinan datasets (Jinan-1, Jinan-2, Jinan-3), using a shared random exploration dataset. The model is saved after each episode and evaluated in real-traffic scenarios. Without masking, the model exhibits erratic behavior, resulting in high ATT values without convergence. We further compare two masking strategies: Gaussian noise masking and fixed-value masking, where the latter includes two experimental setups—zero-masking and mean-based masking. Zero-masking fails to converge similarly to the unmasked case, while mean-based masking shows a partial convergence trend. However, both fixed-value masking methods are significantly outperformed by our proposed Gaussian noise masking, which achieves superior optimization. This comparison underscores the critical role of Gaussian noise masking in enhancing the performance of our PH-DDPG algorithm.

\subsection{Parameter Sensitivity Analysis}

The mini-batch size $ N $ for Gaussian exploration noise significantly impacts the training performance of our model. To quantify this effect, we conducted four experiments for each $ N $ value, with each experiment limited to a maximum of 80 episodes. Table~\ref{tab:N_sensitivity} summarizes the results across three datasets (Jinan-1, Jinan-2, and Jinan-3), reporting the average travel time (ATT) and the mean episode at which the optimal ATT was achieved.

As shown in Table~\ref{tab:N_sensitivity}, larger mini-batch sizes $ N $ generally yield superior ATT performance, with $ N = 250 $ achieving the lowest ATT values across all datasets. This improvement is attributed to the increased diversity of Gaussian noise samples, which enhances the stability of gradient estimation during training. However, the benefits of larger $ N $ must be balanced against computational efficiency and convergence speed. While $ N = 250 $ achieves the best theoretical ATT performance, it requires significantly more episodes to converge compared to smaller batch sizes.

For a deeper understanding of this phenomenon, we analyze the gradient contribution using the following mathematical framework. The gradient of each reward $ r $ with respect to the action parameters $ x $ is computed as a proportion of the total gradient of the sum of rewards $ \sum_{k \in [K]} r $. Specifically, the gradient contribution matrix is derived from the Jacobian matrices of $ \boldsymbol{r} $ with respect to $ \boldsymbol{x} $:

\begin{equation}
    {Contr}_{ij} = \left| \sum_{n=1}^{N} \left( \frac{\partial r_i^{(n)}}{\partial x_j^{(n)}} \Big/ \frac{\partial \sum_{k \in [K]} r_k^{(n)}}{\partial x_j^{(n)}} \right) \right|.
\end{equation}

Each row of the gradient contribution matrix is then normalized proportionally:

\begin{equation}
    \tilde{Contr}_{ij} = \frac{Contr_{ij}}{\sum_{i=1}^{K} Contr_{ij}}.
\end{equation}

\begin{figure*}[ht]
	\centering
	\includegraphics[width=0.8\textwidth]{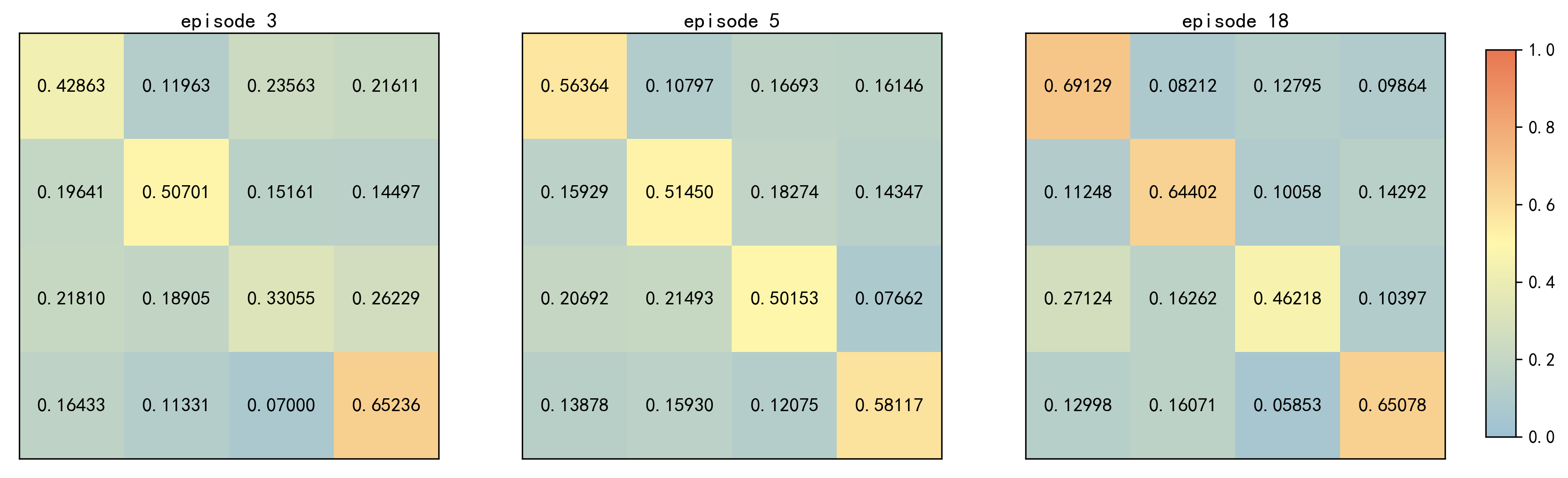}
	\caption{The normalized proportional gradient contribution matrices from episodes 3, 5, and 18, generated during the training of PH-DDPG on the Jinan-1 datasets with $N = 80$.}
	\label{fig_gpn80}
\end{figure*}

\begin{figure*}[ht]
	\centering
	\includegraphics[width=0.8\textwidth]{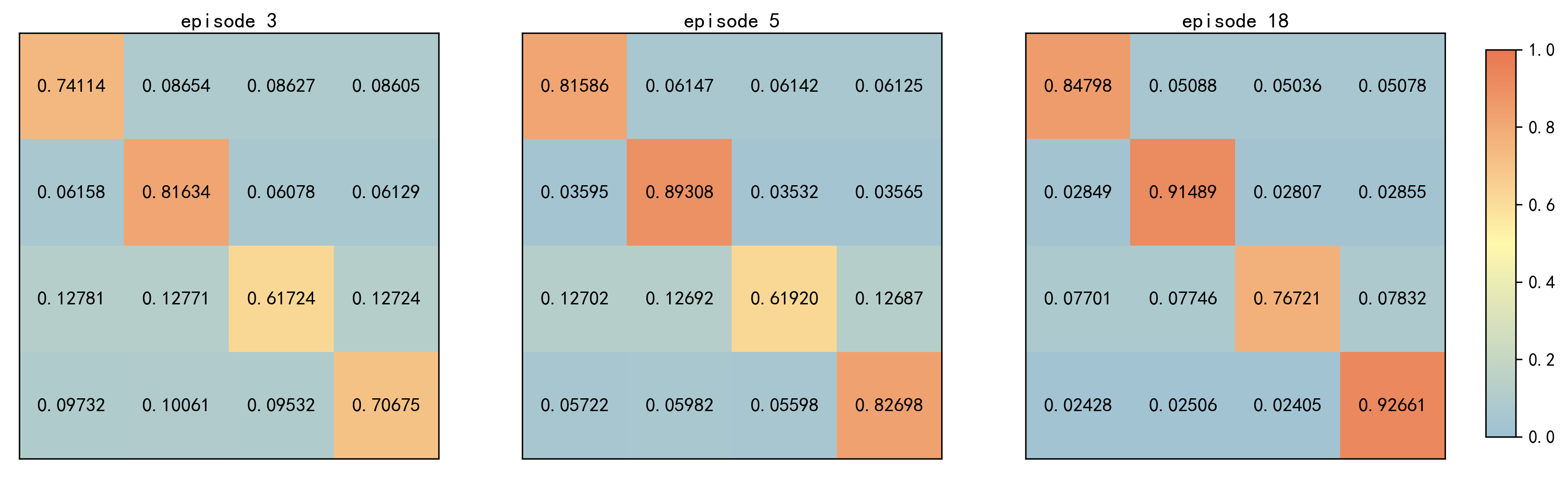}
	\caption{The normalized proportional gradient contribution matrices from episodes 3, 5, and 18, generated during the training of PH-DDPG on the Jinan-1 datasets with $N = 250$.}
	\label{fig_gpn250}
\end{figure*}

Visualizations of $\tilde{Contr}_{ij}$ in Figures~\ref{fig_gpn80} and \ref{fig_gpn250} analyze the gradient patterns extracted during Actor network training, specifically investigating the average directional influence of action rewards on their corresponding parameters. The observed diagonal dominance in these matrices reveals fundamental relationships: higher diagonal intensity indicates that each action's Q-value prediction predominantly relies on its own parametric configuration ($x_i^{(n)}$ for action $a_i$), while off-diagonal elements quantify the unintended coupling where Q-estimates become influenced by extraneous action parameters ($x_j^{(n)}$ where $j \neq i$). The strengthened diagonal pattern with larger $N$ (clearly seen in Fig.~\ref{fig_gpn250}) confirms that increasing $ N $ enhances the effectiveness of the mask function in minimizing noise relative to the total gradient, thereby stabilizing the training process.

% Visualizations of $ \tilde{Contr}_{ij} $ in Figures \ref{fig_gpn80} and \ref{fig_gpn250} demonstrate that larger mini-batch sizes $ N $ result in a more pronounced diagonal pattern. This indicates that increasing $ N $ enhances the effectiveness of the mask function in minimizing noise relative to the total gradient, thereby stabilizing the training process.

In practice, we select $N = 80$ as the optimal mini-batch size for both online and offline training modes. This choice balances computational efficiency with performance, particularly in online learning scenarios where rapid convergence is critical. Although larger $N$ values slightly improve ATT, they require additional training iterations or larger training sets, which is undesirable in time-sensitive applications.

% \subsection{Hyperparameter sensitivity}

% For gradient analysis, we calculate the gradient of each reward $r$ with respect to each action parameter $x$ as a proportion of the gradient of the sum of all rewards $\sum_{k \in [K]} r$. The gradient contribution matrix is computed using the Jacobian matrices of $\boldsymbol{r}$ with respect to $\boldsymbol{x}$, applying the formula:
% \begin{equation}
%     {Contr}_{ij} = \left| \sum_{n=1}^{N} \left( \frac{\partial r_i^{(n)}}{\partial x_j^{(n)}} / \frac{\partial \sum_{k \in [K]} r_k^{(n)}}{\partial x_j^{(n)}} \right) \right|.
% \end{equation}
% Each row of the gradient contribution matrix is then normalized proportionally:
% \begin{equation}
%     \tilde{Contr}_{ij} = \frac{Contr_{ij}}{\sum_{i=1}^{K} Contr_{ij}}.
% \end{equation}

% Visualizations of $\tilde{Contr}_{ij}$ reveal distinct patterns influenced by different replay buffer capacities, as shown in Figures \ref{fig_gpn80} and \ref{fig_gpn250}. Larger sample sizes result in a more pronounced diagonal pattern, indicating the effectiveness of the mask function in minimizing noise relative to the total gradient.

\section{Conclusion}

This study introduces PH-DDPG, a unified reinforcement learning framework that simultaneously optimizes traffic signal phase sequencing and duration through integrated parameter learning—contrasting with conventional two-stage optimization approaches—thereby improving real-world applicability. To address decoupled action constraints in dynamic environments, we designed a Gaussian Noise based action masking mechanism. The framework’s dual offline/online deployment modes ensure operational flexibility, while extensive empirical evaluations on real-world traffic datasets demonstrate significant superiority over state-of-the-art baselines. Also, we propose transitional algorithm variants to facilitate practical adoption, enabling incremental migration from rule-based configurations to fully adaptive control through configurable action boundaries—a critical feature for balancing stability and adaptability in industrial deployments. Nevertheless, limitations persist: our evaluation scenarios are predominantly confined to established urban traffic patterns, lacking systematic assessment in alternative hybrid action space decision-making scenarios. Future investigations prioritize two critical extensions:(1) advancing hybrid action space decision-making through multi-agent coordination mechanisms specifically designed for large-scale urban networks, and (2) systematically extending the application scenarios of this methodologies in hybrid action space settings.

\bibliographystyle{unsrt} 
\bibliography{main}

% \begin{IEEEbiography}[{\includegraphics[width=1in,height=1.25in,clip,keepaspectratio]{fig1.png}}]{IEEE Publications Technology Team}
% In this paragraph you can place your educational, professional background and research and other interests.\end{IEEEbiography}

\end{document}